\documentclass[conference]{IEEEtran}
\IEEEoverridecommandlockouts
\usepackage{cite}
\usepackage{amsmath,amssymb,amsfonts}
\usepackage{algorithmic}
\usepackage{graphicx}
\usepackage{textcomp}
\usepackage{xcolor}
\usepackage{subcaption}

\usepackage{tabu}                     
\usepackage{multirow}                 
\usepackage{multicol}                 
\usepackage{multirow}                
\usepackage{float}                    
\usepackage{makecell}                 
\usepackage{booktabs}                 

\def\BibTeX{{\rm B\kern-.05em{\sc i\kern-.025em b}\kern-.08em
    T\kern-.1667em\lower.7ex\hbox{E}\kern-.125emX}}

\begin{document}

\title{ Enhancing Shape Perception and Segmentation Consistency for Industrial Image Inspection\\
\thanks{This work is supported by Chinese National Natural Science Foundation under Grants 62076033. \par $^*$corresponding author \par $^\dagger$Equal contribution}
}


\author{\IEEEauthorblockN{Guoxuan Mao$^{1,\dagger}$,
Ting Cao$^{2,\dagger}$,
Ziyang Li$^{1,\dagger}$,
Yuan Dong$^{1,*}$}
\IEEEauthorblockA{$^1$School of Artificial Intelligence,
Beijing University of Posts and Telecommunications, Beijing 100876
, China}
\IEEEauthorblockA{$^2$Ricoh Software Research Center (Beijing) Co., Ltd, Beijing, China}
\IEEEauthorblockA{m\_3475@bupt.edu.cn, ting.cao@cn.ricoh.com, liziyang990531@foxmail.com, yuandong@bupt.edu.cn}
}

\maketitle

\begin{abstract}

Semantic segmentation stands as a pivotal research focus in computer vision. In the context of industrial image inspection, conventional semantic segmentation models fail to maintain the segmentation consistency of fixed components across varying contextual environments due to a lack of perception of object contours. Given the real-time constraints and limited computing capability of industrial image detection machines, it is also necessary to create efficient models to reduce computational complexity. In this work, a Shape-Aware Efficient Network (SPENet) is proposed, which focuses on the shapes of objects to achieve excellent segmentation consistency by separately supervising the extraction of boundary and body information from images. In SPENet, a novel method is introduced for describing fuzzy boundaries to better adapt to real-world scenarios named Variable Boundary Domain (VBD). Additionally, a new metric, Consistency Mean Square Error(CMSE), is proposed to measure segmentation consistency for fixed components. Our approach attains the best segmentation accuracy and competitive speed on our dataset, showcasing significant advantages in CMSE among numerous state-of-the-art real-time segmentation networks, achieving a reduction of over $\pmb{50\%}$ compared to the previously top-performing models.
\end{abstract}


\section{Introduction}
Semantic segmentation stands as one of the most pivotal tasks in the field of computer vision. In the realms of medicine, industry, and autonomous driving, it has a wide range of applications. Our research concentrates on specific industrial scenarios characterized by a limited number of segmentation categories. 
\par In recent years, with the advancement of deep learning, semantic segmentation has continuously made breakthroughs in accuracy, particularly when incorporating with Transformers \cite{vaswani2017attention}. After pre-training on large-scale datasets, these models have demonstrated performance surpassing traditional convolutional neural networks. However, models that achieve high accuracy typically accompany larger parameter sizes, increased computational resource demands, and lower inference efficiency. In the industrial scenarios that necessitate semantic segmentation techniques, such as defect detection and quality inspection of industrial products, there is a demand for rapid quality assessment of large batches of products within a short timeframe. Therefore, the selected model must prioritize efficiency while ensuring accuracy to meet the demands of real-time quality assessment in industrial settings.
\par Taking the PVC coating dataset of the vehicle undercarriage, which will be detailed in \ref{sec4-a} and is the main focus of this paper, as an example, images in industrial scenarios for inspection typically exhibit the following characteristics: (1) In standardized production, the shooting position of the target area for inspection remains consistent.  (2) The target categories requiring pixel-wise classification are relatively few. (3) In various images, the relative positions and sizes of identical components exhibit general consistency. The characteristic shapes of certain components remain fixed, necessitating the model to accomplish more accurate and consistent segmentation, particularly for smaller components that pose challenges in segmentation. 
\par Similar to many real-time semantic segmentation networks, we have also proposed a dual-path structure. In the semantic path, we leverage widely recognized backbones such as ResNet \cite{he2016deep}. The primary role of our spatial path is to extract shape information. We consistently maintain the feature map in a high-resolution state. Generally, the edges of objects arise from differences between nearby pixels which is a local feature, and there is not a strong reliance on spatial relationships across larger contexts. Hence, in our spatial path, we opt for the use of asymmetric convolutions. On the one hand, this reduces parameters under similar receptive fields, and on the other hand, a 1×n convolution kernel can better match some edge features. Additionally, we employ dilated convolutions \cite{yu2017dilated} with varying dilation rates to further increase the receptive field. As the image undergoes downsampling, we reduce the maximum kernel size of asymmetric convolutions and the maximum dilation rate of dilated convolutions to decrease parameters. Simultaneously, drawing inspiration from \cite{li2020improving}, we devised a flow-based module named decoupled module to separate high-frequency and low-frequency information in features. Regarding supervision, we designed a more rational approach named Variable Boundary Domain to determine whether a pixel belongs to the boundary. Compared to the original method, our proposed approach achieved superior results.

\begin{figure*}[htbp]
\centerline{\includegraphics[width=0.95\textwidth,keepaspectratio]{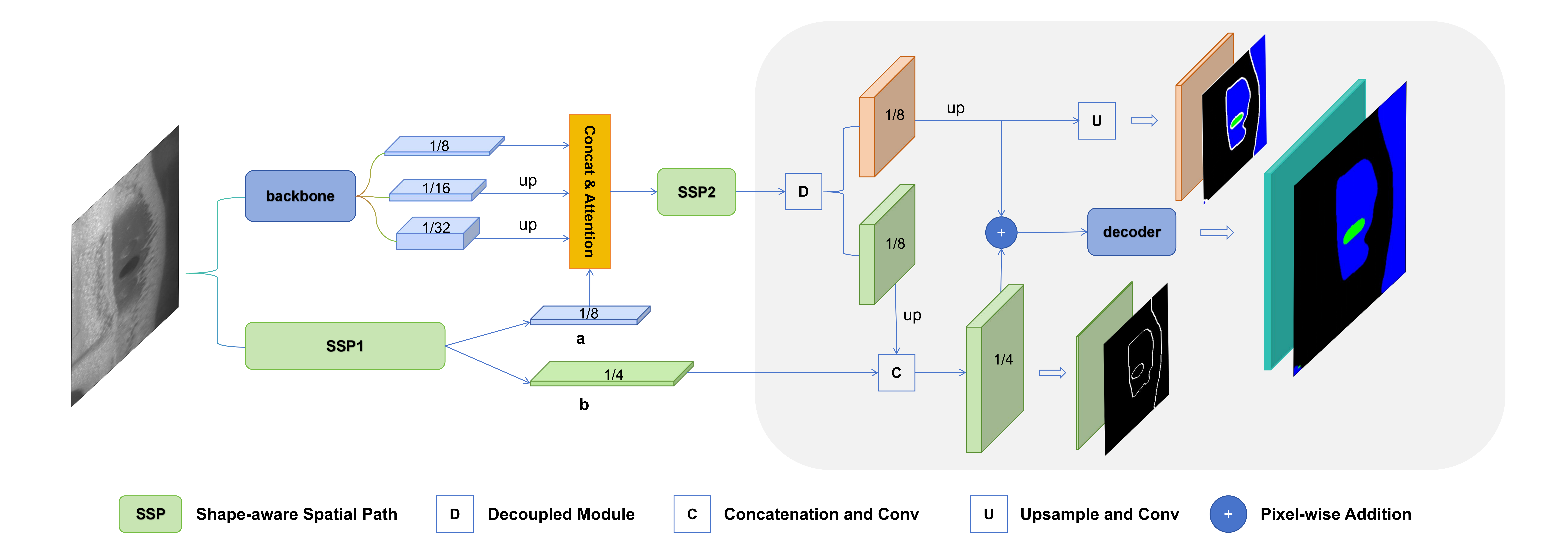}}
\caption{Overall structure of our SPENet, the detailed information of "SSP" is shown in Fig. \ref{fig2}. We annotated the resolution size of the critical intermediate features relative to the input image. The "Decoupled Module" is a method proposed in \cite{li2020improving} for separating body and edge information.}
\label{fig1}
\end{figure*}

\par In semantic segmentation, the most commonly used evaluation metric is mIoU, which calculates the overlap between predicted and ground truth results. However, for industrial images, certain fixed components require more consistent segmentation. We observed that when using certain models, the segmentation shapes of fixed components vary with different contexts, which probably impacts post-segmentation tasks such as handling distances between different parts. Therefore, we propose a new metric, Consistency Mean Square Error(CMSE), to assess the segmentation consistency for the same component. A lower CMSE value indicates higher consistency. The CMSE metric of our model demonstrates significant advantages among numerous models.
\par The proposed SPENet with VBD achieves $95.88\%$ mIoU at 81 FPS on the PVC dataset with the GTX 3060, and  $27.09 \times 10^{-4}$ CMSE, which significantly surpasses other state-of-the-art real-time networks such as BiseNet\cite{yu2018bisenet} and DDRNet\cite{hong2021deep}. On classic public datasets such as Cityscapes, it also demonstrates competitive performance.

\section{Related Work}
\subsection{Semantic segmentation}
Recent approaches in Semantic Segmentation have shown a notable development trend in many directions. Early work in this field is almost based on convolutional neural networks and FCNs \cite{long2015fully} are widely adopted with the encoder-decoder architecture. Unet \cite{ronneberger2015u} pioneered a skipped-layer architecture to integrate features at different scales and this idea is reflected frequently in the following studies. The pyramid pooling module (PPM) in PSPNet \cite{zhao2017pyramid} and the Atrous Spatial Pyramid Pooling (ASPP) in DeepLab v3 \cite{chen2018encoder} are also proven to be effective methods in capturing multi-scale contexts. After the transformer \cite{vaswani2017attention}was introduced in computer vision which was at first designed for nlp, the Performance in various vision tasks has improved significantly depending on its Strong encoder based on multi-head self-attention mechanisms, especially with a large scale of data accessed. However, the Excellent Accuracy is obtained at the expense of memory consumption and inference efficiency, which makes it difficult to deploy on inspection machines in the field of Industry. Our model is a light-weighted CNN which is memory-saving and efficient. 
\subsection{Real-time semantic segmentation CNN Architectures  }
Depth-wise separable convolution is the prevalent technique employed in real-time semantic segmentation models such as ESPNet \cite{mehta2018espnet}. FastSCNN \cite{poudel2019fast} designs a two-stream architecture, one of which focuses on deep semantic information with low-resolution inputs while the other focuses more on Spatial details with high-resolution inputs. Its lightweight design ensures high efficiency, but the accuracy cannot be guaranteed. DDRNet \cite{hong2021deep} similarly adopts a dual-path design, with continuous feature interaction between the two paths during the forward process, showcasing both fast and accurate. Nevertheless, during the output segmentation, DDRNet utilizes a direct 8x upsampling approach, resulting in prominent jagged artifacts, particularly in low-resolution images. Our SPENet adopts a smoother upsampling approach, allowing the output results to better preserve fine details.
\subsection{Semantic segmentation focusing on the boundary }
Generally, the accuracy of boundary segmentation is a challenging problem due to the apparent uncertainty of pixel classification When transitioning from one object to another. To tackle this issue, Lee et al. \cite{lee2020structure}Proposed a structure boundary preserving framework that reinforces boundary information by Key point Map Generator; Gated-SCNN \cite{takikawa2019gated} explored a shape stream with a shallow architecture which is allowed to only focus on the edge information and operates on full image resolution. Li et al. proposed the decoupled module \cite{li2020improving} that obtains body feature by sampling from a lower-resolution feature map, the residual between the body feature and original feature is denoted as the edge feature, then both of them are supervised by the corresponding masks separated from the intact ground truth. But all the feature information is learned by the backbone ResNet whose ability is limited for Separating edge and body features. Drawing inspiration from this approach, we formulated a boundary domain supervision method and incorporated this module into a lighter-weight and two-stream network, yielding favorable results on prevalent industrial datasets.

\section{Methods}
\subsection{Spatial Path for Shape Extraction \label{sec a}}
\begin{figure}[htbp]
\centerline{\includegraphics[width=0.48\textwidth,keepaspectratio]{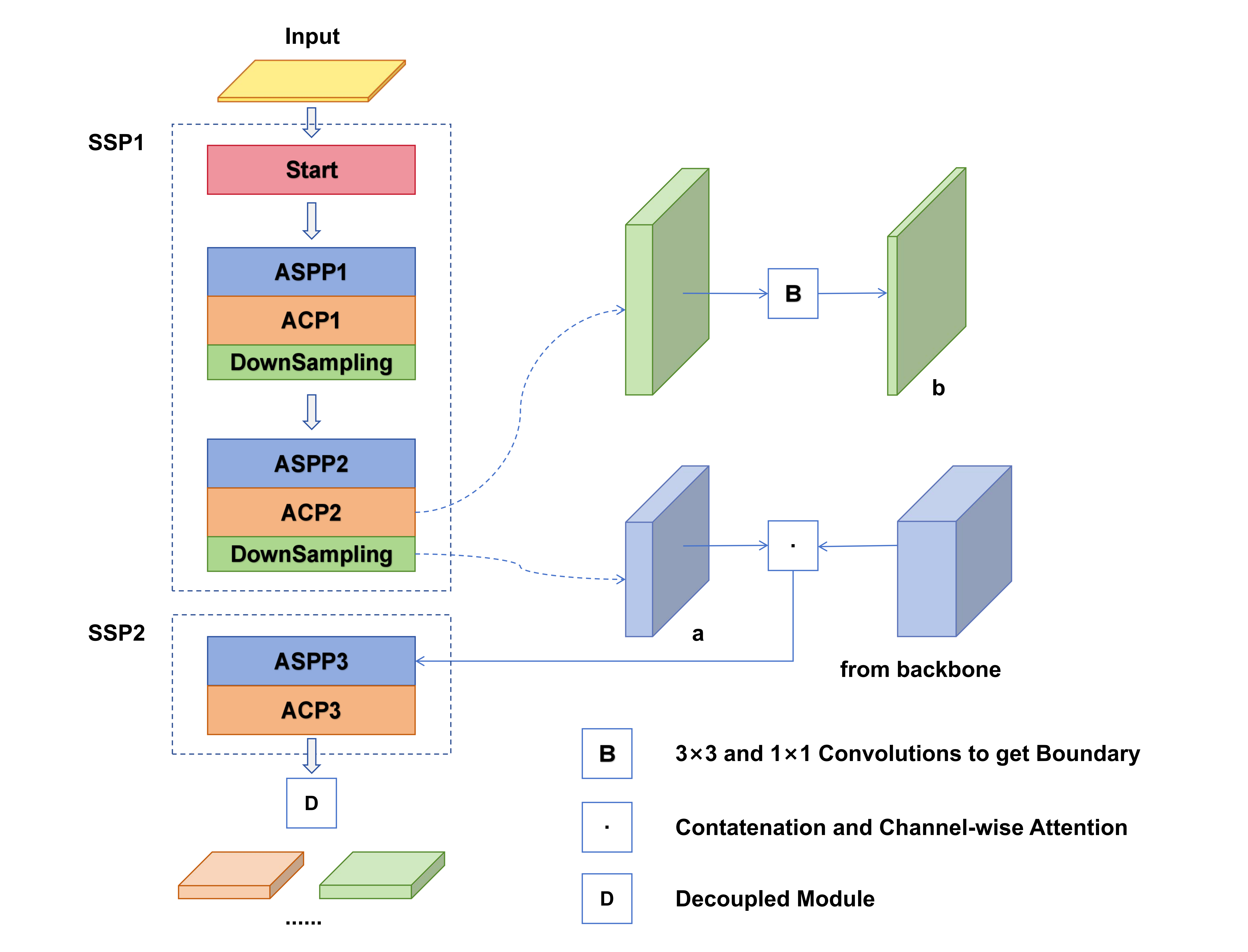}}
\caption{The architecture of "SSP" in Fig. \ref{fig1}. Channel-wise Attention is the Squeeze-and-Excitation process in \cite{hu2018squeeze}. "a" and "b" correspond to the "a" and "b" in Fig. \ref{fig1}. The "Start" utilizes convolution with the stride of 2 combined with MaxPooling in \cite{paszke2016enet}.}
\label{fig2}
\end{figure}

With an RGB input image $I \in \mathbb{R}^{3\times H \times W }$, we first use a module to downsample $I$ to the resolution 1/2 that concatenates the feature map from convolution operation with the stride of 2 and max-pooling like ENet\cite{paszke2016enet} to get the feature map $F \in \mathbf{R}^{C\times H/2 \times W/2 } $. Then we put the feature map F into the following blocks including ASPP and ACP in Fig \ref{fig2}. \par
\begin{figure*}[htbp]
\centerline{\includegraphics[width=0.95\textwidth,keepaspectratio]{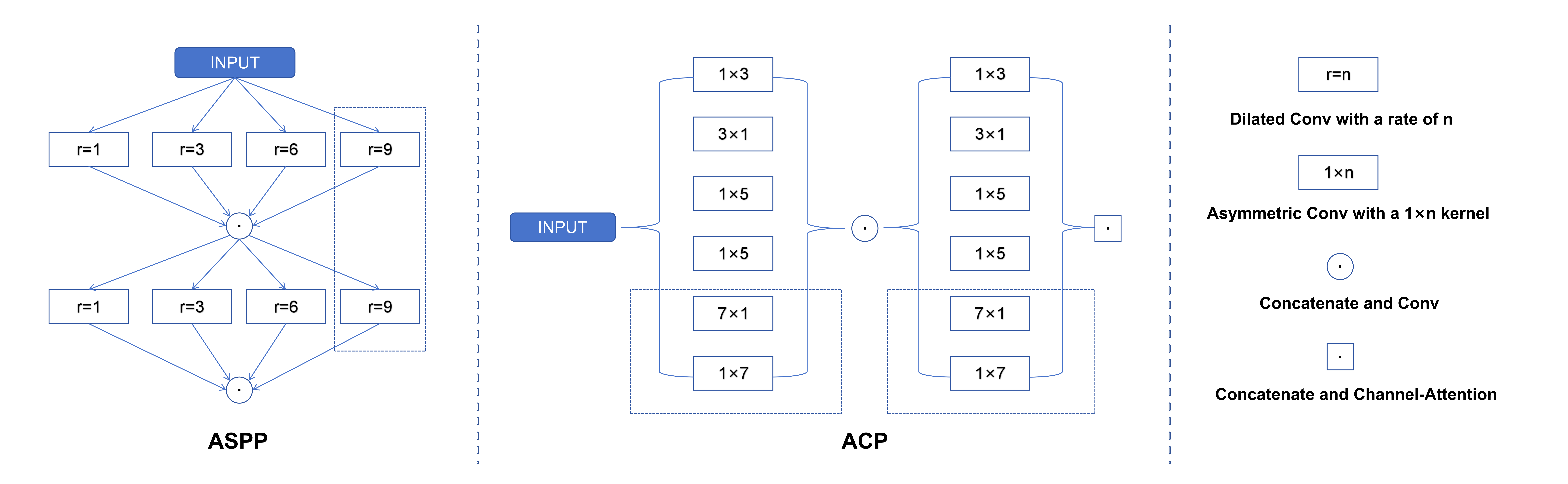}}
\caption{The detail of ASPP and ACP in Fig \ref{fig2}, all the dilated and asymmetry convolutions are depth-wise separable convolutions.}
\label{fig3}
\end{figure*}

 ASPP is a classic module proposed in DeepLab\cite{chen2017deeplab} which is used to enlarge the receptive field of the fixed-kernel-size convolution, we made some adjustments shown in Fig \ref{fig3}.  We set a series of dilated ratio $\left \{  1, 3, 6, 9\right \}$, in ASPP2 and ASPP3 in Fig \ref{fig2} we get rid of $r = 9$ to reduce the consumption of memory and calculate because after downsampling the excessively large receptive field is not necessary for this shape focusing path. ACP is composed of different ACs(asymmetric convolutions) whose construction and kernel size are shown in Fig \ref{fig3}. ACs are also a strategy to reduce parameters and can fit the slender shape of the boundary. When the feature map is downsampled to 1/8-resolution, We adopted two different blocks: ASPP3 and ACP3 to produce the semantic feature, and $1\times 1$ convolution to produce the boundary feature shown in Fig \ref{fig2}, the two feature maps will be employed in the next stage. To reduce the parameters, depth-wise convolution is applied in the ASPP and ACP blocks. After every ACP block, we use a normal Conv as the bottleneck to merge the different features and reduce the channels after concatenation. Then the channel-wise attention\cite{hu2018squeeze} is utilized for optimizing the features.

\subsection{Shape-aware efficient network}\label{sec b}
We design a two-path network, the Spatial path is introduced in Sec A, Another path is a backbone for extracting deep semantic information. We utilize resnet-18\cite{he2016deep} with half of the channels compared to the standard model considering our task of small-class segmentation for industrial images. the input image is progressively downsampled through the backbone to the resolution of $\left \{  1/2,1/4,1/8,1/16,1/32\right \}$, during the forward propagation we concatenate the last 1/8-resolution feature map, the interpolated 1/16, 1/32-resolution feature and the semantic feature from the spatial path, then employ Conv to obtain a feature of global multi-level semantic information. Following this we adopt the decoupled module\cite{li2020improving} which proposes an upsampling strategy by the flow of the feature to get the feature with more precise body information and more ambiguous boundary information. This operation draws on the idea of Gaussian filtering for smoothing through which the high-frequency boundary feature can be obtained by subtraction. The final boundary feature is processed as a single channel form for supervision. The decoupled body feature and boundary features are then added by pixels. Finally, we employ a two-stage decoder: two rounds of upsampling followed by convolution to obtain the segmentation results at the original resolution.

\subsection{Variable Boundary Domain}
In section \ref{sec b} there are three parts of outputs from our network: body, boundary, and integral segmentation map, all of which need to be separately supervised. For the integral part we use conventional Focal Loss and Dice Loss,  For the remaining two parts, we extract the binary boundary masks from the full mask. When calculating the loss of the body part, we ignore the impact pixel location of the valid pixels from the boundary mask. 
Furthermore, We propose a novel approach for describing the edge named Variable Boundary Domain(VBD). We extract the edge mask based on the distance from different class regions. In \cite{li2020improving} different distance thresholds are used to determine whether a pixel is a boundary or not. But in most instances, it is quite difficult to precisely distinguish the discrete boundary just by one or two continuous pixels for the unavoidable noise. Hence, We apply the Gaussian function about the distance to build a boundary probability map $p \in \mathbb{R}^{H \times W}$. $p_{i}$ denotes the probability of pixel $i$ being the boundary and is calculated by the formulation shown in Equation \ref{eq1}. 
\begin{equation}
p_{i} =  e^{-\frac{(d_{i} - 1)^2}2}\label{eq1}
\end{equation}
\par The $d_{i}$ denotes the distance from pixel $i$ to the nearest pixel of the other classes. Based on the probability domain we generate different boundary masks $e$ for every training data. The boundary loss is calculated by Weighted Binary Cross-Entropy (BCE) Loss as shown in Equation \ref{eq2}. The body loss is calculated by Cross-Entropy(CE) Loss and Dice Loss as shown in Equation \ref{eq3}
\begin{equation}
    L_{e}(y_e,\hat{y_e}) = - w \cdot (\hat{y_{e}} \cdot \log(y_{e}) - (1 - \hat{y_{e}}) \cdot \log(1 - y_{e}))\label{eq2}
\end{equation}

 The $y_e$ and $y_b$ denote the output of the edge feature and body feature while $\hat{y_e}$ and $\hat{y_b}$ denote their masks. $w \in \mathbb{R}^{H \times W}$ is employed to balance the disparity in the number of samples between boundary and non-boundary regions.

\begin{equation}
    L_b(y_b,\hat{y_b} ) = \lambda \cdot L_{ce}(y_b,\hat{y_b}) + (1-\lambda) \cdot   L_{dice}(y_b,\hat{y_b})\label{eq3}
\end{equation}
The final output of our SPENet is supervised by the ground-truth  $\hat{s}$ applying the Focal loss and Dice loss. We eventually sum the four components with different weights $\lambda$ which is shown in Equation \ref{eq4}. All default values for $\lambda$ are set to $0.5$.
\begin{equation}
    L = \lambda_{1}\cdot L_e+\lambda _{2}\cdot L_b+\lambda_3\cdot L_{focal}(s,\hat{s} )+\lambda _4\cdot L_{dice}(s,\hat{s} )\label{eq4}
\end{equation}

\subsection{Consistence Stability metrics for Segmentation}
We proposed a novel metric: Consistency Mean Square Error(CMSE) to estimate the segmentation performance from the aspect of segmentation consistency(SC) of a fixed component. We take the location-hole from the PVC test dataset as an example,  we crop it from the segmentation map by a rectangular box that can just completely cover it. This cropped patch is resized to the same size and converted into binary form. We calculate the average segmentation result for these patches through a statistical approach: initially, we compute the average pixel number of the location hole across all patches denoted by $n$. Subsequently, we sum all the binary patches to obtain $t \in \mathbb{R}^{H \times W}$ and select $n$ pixels with the highest values as the binary average result $\bar{m} \in \mathbb{R}^{H \times W}$ of the holes. We calculate the Intersection over Union($IoU$) between all patches and $\bar{m}$ and define $1-IoU$ as the error of SC. The CMSE is defined as shown in Equation \ref{eq5}. $N$ denotes the number of images in the test dataset.

\begin{equation}
    CMSE = \frac{1}{N}\cdot \sum_{i=1}^{N} {(1-IoU(M_i,\bar{m}))^2} \label{eq5}
\end{equation}



\section{Experiment}

\subsection{Datasets}\label{sec4-a}

\begin{figure}[htbp]
\centering
  \begin{subfigure}{0.22\linewidth}
    \includegraphics[width=\linewidth]{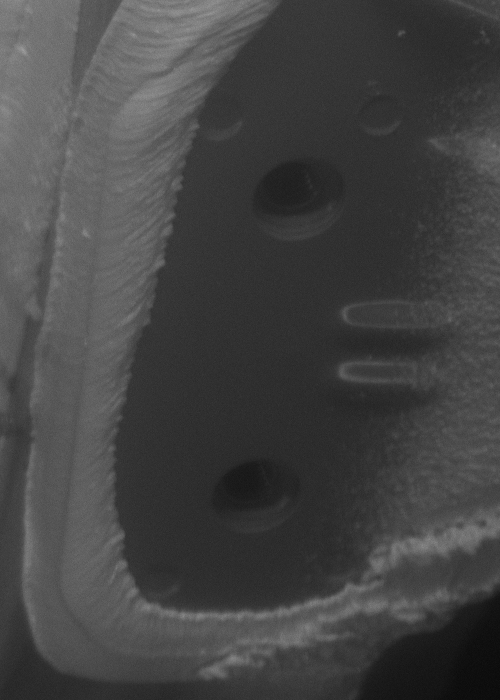}
    \label{fig4:sub1}
  \end{subfigure}%
  \begin{subfigure}{0.22\linewidth}
    \includegraphics[width=\linewidth]{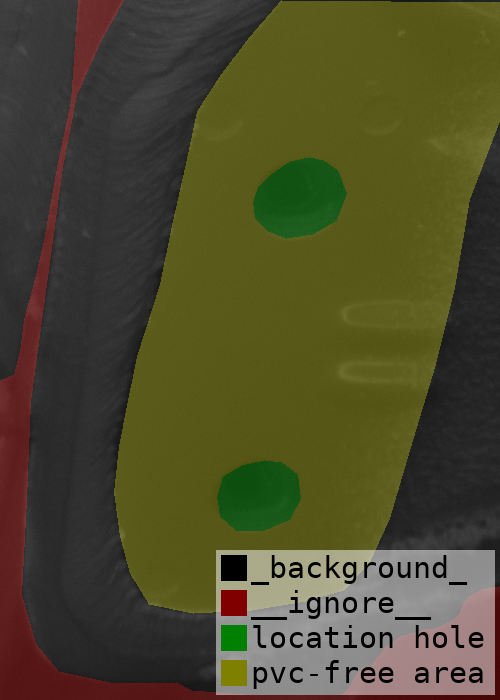}
    \label{fig4:sub2}
  \end{subfigure}
  \begin{subfigure}{0.22\linewidth}
    \includegraphics[width=\linewidth]{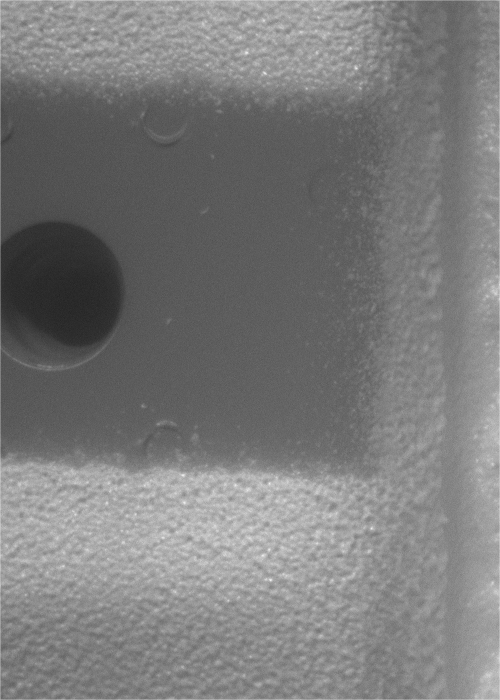}
    \label{fig4:sub3}
  \end{subfigure}%
  \begin{subfigure}{0.22\linewidth}
    \includegraphics[width=\linewidth]{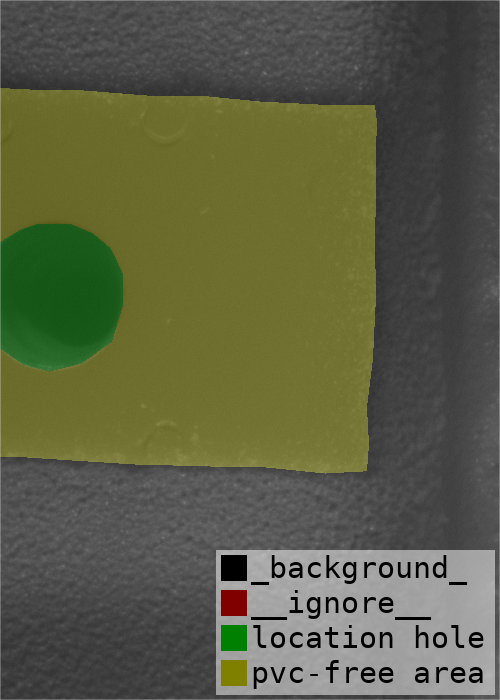}
    \label{fig4:sub4}
  \end{subfigure}

  \caption{Example of PVC dataset}
  \label{fig4}
\end{figure}

In this paper, we focus on industrial images with similar hue, lightness, and saturation which require applying semantic segmentation methods for quality inspection. Our thinking originates in the PVC datasets, thus we choose it as the main dataset for our research. The PVC dataset was captured from the underside of the vehicles and contains 2 types of vehicles and 27 locations in total. The PVC coating on the bottom of vehicles is a crucial process to ensure the sealing, dust-proofing, noise reduction, and corrosion resistance of the vehicle underbody. Our target is to accomplish the pixel-wise classifier of the three objects: PVC coating(background), PVC-free area, and location hole, which are shown in Fig \ref{fig4}, the relative positioning of PVC and location hole serves as the basis for our assessment of whether the coating placement is satisfactory. There are 500 image-label pairs for training evenly distributed across each location of the vehicle except one we designate as the test set to validate the network's generalization capability.
\par Besides, to demonstrate the effectiveness of the proposed method on a broader range of data, we also trained our model with CityScapes and compared our results with other classic semantic segmentation models trained under the same setup and conditions. 

\subsection{Implementation details}
For both PVC and CityScapes, we use the stochastic gradient descent (SGD) optimizer with an initial learning rate of 0.01, a momentum of 0.9, and a weight decay of 0.0005. We apply a polynomial decay policy with the power of 0.9 to drop the learning rate. Random horizontal flip, histogram equalization, brightness and contrast jittering, as well as random Gaussian noise are adopted for data augmentation. For PVC, images are resized to 448×448 since the size and aspect ratio of images collected from different positions vary significantly. For CityScapes, images are resized to 512×256 for both training and validation. All the experiment is conducted on a single GPU(NVIDIA GeForce RTX 4090) with a batch size of 12. We trained PVC for 300 epochs and CityScapes for 250 epochs.

\begin{figure*}[h]
 \centering
	
 \begin{minipage}{0.15\linewidth}
 	\vspace{3pt}
 	\centerline{\includegraphics[width=\textwidth]{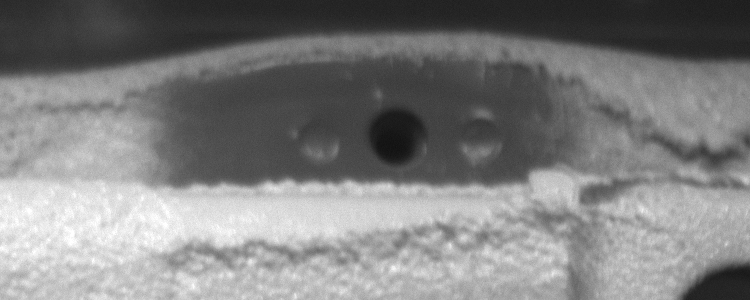}}
 	\vspace{3pt}
 	\centerline{\includegraphics[width=\textwidth]{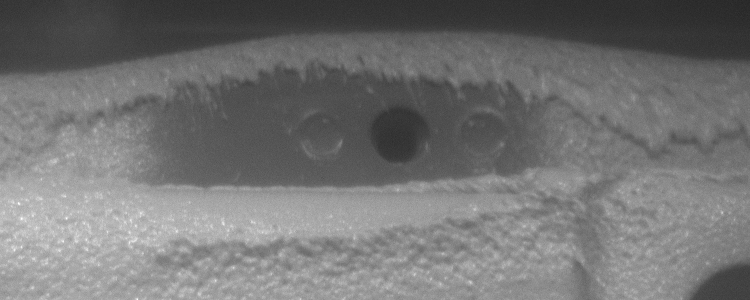}}
 	\vspace{3pt}
 	\centerline{Image}
 \end{minipage}
 \begin{minipage}{0.15\linewidth}
 	\vspace{3pt}
 	\centerline{\includegraphics[width=\textwidth]{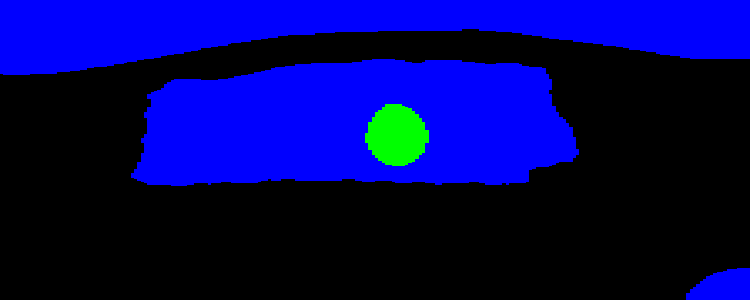}}
 	\vspace{3pt}
 	\centerline{\includegraphics[width=\textwidth]{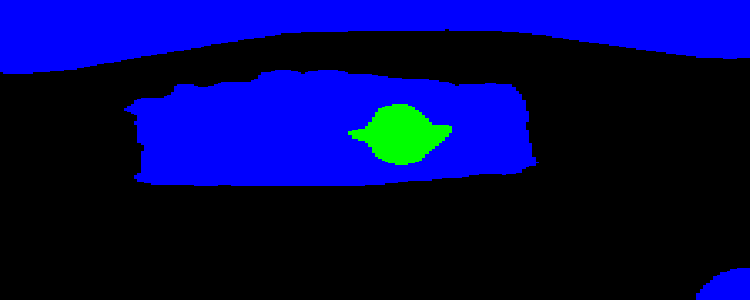}}
 	\vspace{3pt}
 	\centerline{UNet}
  \end{minipage}
  \begin{minipage}{0.15\linewidth}
 	\vspace{3pt}
 	\centerline{\includegraphics[width=\textwidth]{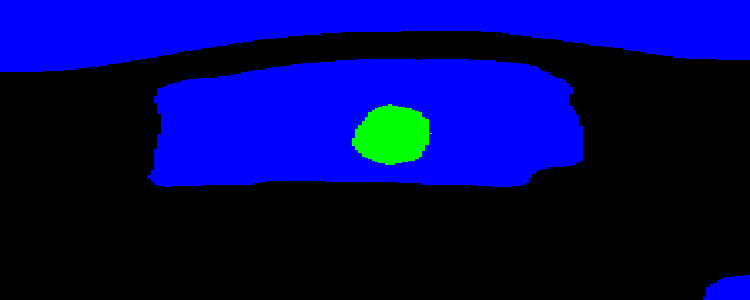}}
 	\vspace{3pt}
 	\centerline{\includegraphics[width=\textwidth]{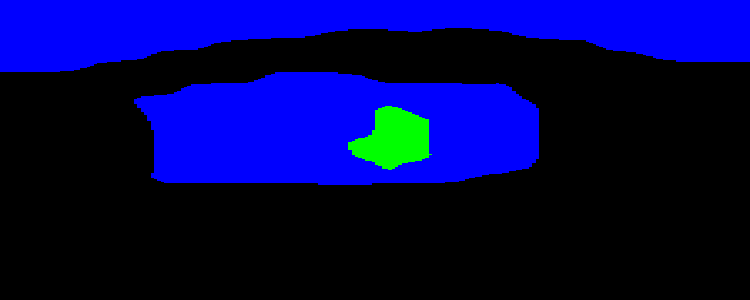}}
 	\vspace{3pt}
 	\centerline{BiseNet}
  \end{minipage}
  \begin{minipage}{0.15\linewidth}
 	\vspace{3pt}
 	\centerline{\includegraphics[width=\textwidth]{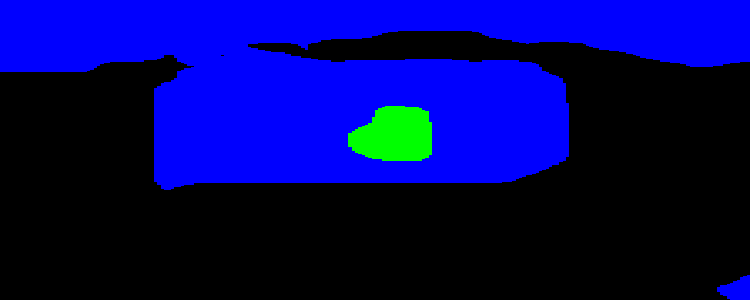}}
 	\vspace{3pt}
 	\centerline{\includegraphics[width=\textwidth]{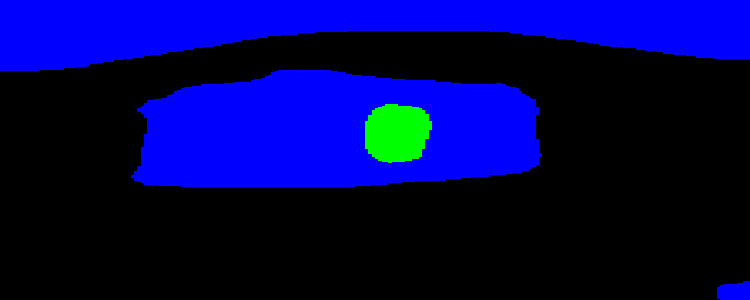}}
 	\vspace{3pt}
 	\centerline{FastSCNN}
  \end{minipage}
  \begin{minipage}{0.15\linewidth}
 	\vspace{3pt}
 	\centerline{\includegraphics[width=\textwidth]{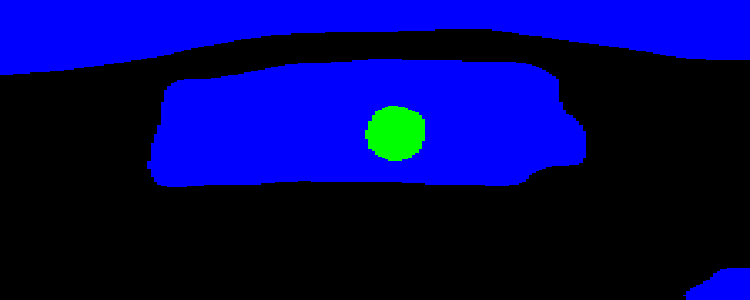}}
 	\vspace{3pt}
 	\centerline{\includegraphics[width=\textwidth]{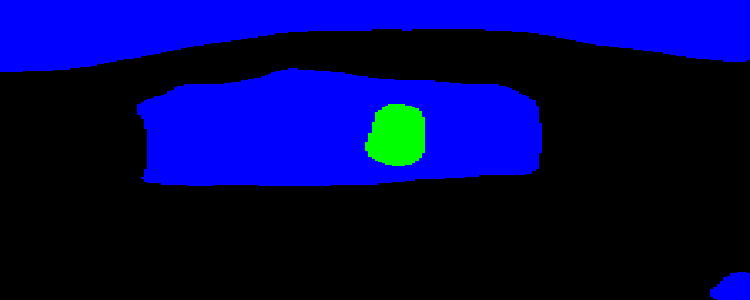}}
 	\vspace{3pt}
 	\centerline{DDRNet}
  \end{minipage}
  \begin{minipage}{0.15\linewidth}
 	\vspace{3pt}
 	\centerline{\includegraphics[width=\textwidth]{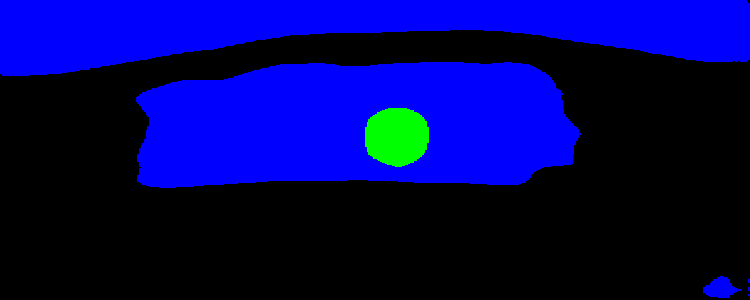}}
 	\vspace{3pt}
 	\centerline{\includegraphics[width=\textwidth]{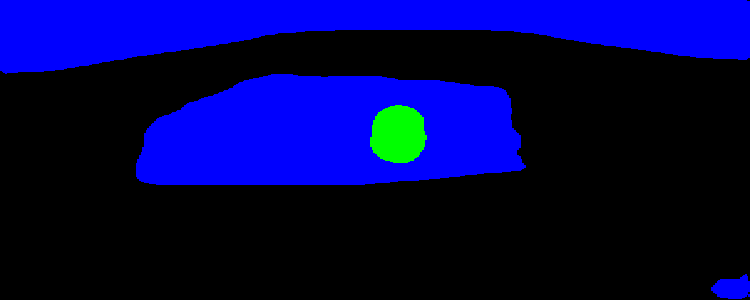}}
 	\vspace{3pt}
 	\centerline{Ours}
  \end{minipage}
  \caption{Visualized segmentation results on PVC dataset}
  \label{fig5}
      
\end{figure*}

\subsection{Accuracy and Efficiency Comparisons}
We evaluate our methods on the PVC validation set and compare the results with state-of-the-art real-time semantic segmentation networks such as FastSCNN, ERFNet, DDRNet, etc. We use pixel-wise accuracy, mean IoU, parameter size(MB), and FPS on a single GPU(Geforce RTX 3060) as the evaluation metrics. As is shown in Table \ref{tab1}, our SPENet outperforms all others in accuracy and mIoU. We outperform the DDRNet, which performs best among models inferior to ours, by an additional $0.37\%$. Although $0.37\%$ is a small difference, it represents a relatively noticeable improvement compared to the differences between other models such as DDRNet, and CGNet. The parameters of SPENet are 3.71 MB, making it smaller than DDRNet and BiseNet, and only slightly larger than FastSCNN and CGNet. The speed of SPENet is better than CGNet and ERFNet but inferior to FastSCNN, DDRNet, and BiseNet, which effectively achieves the requirements for real-time segmentation.

    \begin{table}[htbp]
    \caption{Results on PVC Across Multiple Networks}
        \begin{center}
        \begin{tabular}{c|cccc}
        
        \cline{1-5} 
        \textbf{Model} & \textbf{Accuracy}& \textbf{mIoU}& \textbf{Param} & \textbf{FPS}\\
        \hline
        UNet \cite{ronneberger2015u}& $98.29\%$& $95.40\%$ &$31.04$M & 43\\
        DeepLabv3+\cite{chen2018encoder}& $98.30\%$& $95.45\%$ & - & -\\
        \hline
        ENet \cite{paszke2016enet}& $95.67\%$& $62.38\%$ &$\pmb{0.4}$M & -\\
        ERFNet \cite{romera2017efficient}& $98.16\%$& $95.10\%$ &$20$M & $61$\\
        CGNet \cite{wu2020cgnet}& $98.28\%$& $95.41\%$ &$0.49$M & $54$\\
        FastSCNN \cite{poudel2019fast}& $97.96\%$& $94.56\%$ &$1.1$M & $\pmb{138}$\\
        LEDNet \cite{wang2019lednet}& $95.34\%$& $61.25\%$ & $0.91$M & $52$\\
        BiseNet \cite{yu2018bisenet}& $98.17\%$& $95.07\%$ & $5.8$M & $114$\\
        DDRNet \cite{hong2021deep}& $98.32\%$& $95.51\%$ & $20.1$M & $97$\\
        \hline
        SPENet& $\pmb{98.39\%}$& $\pmb{95.88\%}$ & $3.71$M & $81$ \\
        \hline
        \end{tabular}
        \label{tab1}
        \end{center}
    \end{table}

\par In the CityScapes experiments, given practical hardware and time constraints, we only ensured that all models were trained under the same settings to obtain a relative metric. There is a certain gap between the achieved metric and the publicly reported theoretical results. The results are shown in Table \ref{tab2}. Our SPENet is specifically designed for industrial images with partially stable features, and it demonstrates promising results even when evaluated on the CityScapes dataset. We outperform BiseNet and FastSCNN by over $5\%$, which demonstrates the robustness of our method across different tasks.
\begin{figure}[htbp]
\centering
  \begin{subfigure}{0.295\linewidth}
    \includegraphics[width=\linewidth]{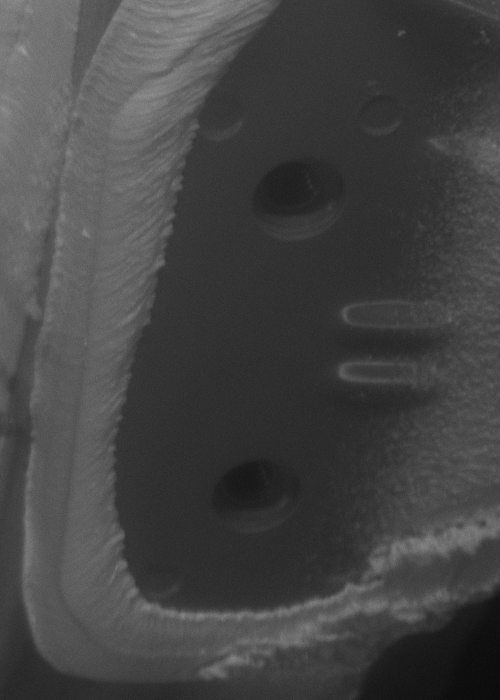}
    \caption{}
    \label{fig:sub1}
  \end{subfigure}
  \begin{subfigure}{0.295\linewidth}
    \includegraphics[width=\linewidth]{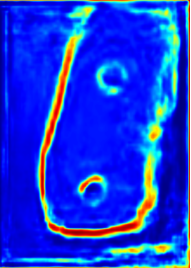}
    \caption{}
    \label{fig:sub2}
  \end{subfigure}
  \begin{subfigure}{0.298\linewidth}
    \includegraphics[width=\linewidth]{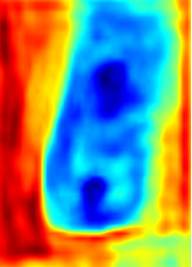}
    \caption{}
    \label{fig:sub3}
  \end{subfigure}
  \caption{Visualization of the decoupled boundary and body map. (a) is the original image, (b) is the boundary map, (c) is the body map }
  \label{fig6}
\end{figure}

    \begin{table}[htbp]
    \caption{Experimental Results on CityScapes}
        \begin{center}
        \begin{tabular}{c|cc}
        
        \cline{1-3} 
        \textbf{Model} &\textbf{Accuracy}& \textbf{mIoU}\\
        \hline
        Fast-SCNN \cite{poudel2019fast}& $89.91\%$& $47.60$  \\
        BiseNet \cite{yu2018bisenet}& $90.59\%$& $50.56\%$  \\
        DDRNet \cite{hong2021deep}& $93.07\%$& $60.67\%$  \\
        \hline 
        SPENet&  $91.75\%$ & $55.73\%$  \\
        \hline
        \multicolumn{3}{l}{These results are a reproduction under specific conditions } \\
        \multicolumn{3}{l}{rather than publicly disclosed outcomes.}
        \end{tabular}
        \label{tab2}
        \end{center}
    \end{table}

\begin{table}[htbp]
    \caption{CMSE Results}
        \begin{center}
        \begin{tabular}{c|cc}
        
        \cline{1-3} 
        \textbf{Model} & \textbf{mIoU}& \textbf{CMSE($1.0\times10^{-4}$)}\\
        \hline
        UNet \cite{ronneberger2015u}& $95.40\%$& $301.48$ \\
        DeepLabv3+\cite{chen2018encoder}& $95.45\%$& $178.05$ \\
        \hline
        ENet \cite{paszke2016enet}& $62.38\%$& -\\
        ERFNet \cite{romera2017efficient}& $95.10\%$& $45.60$  \\
        CGNet \cite{wu2020cgnet}& $95.41\%$& $66.04$ \\
        FastSCNN \cite{poudel2019fast}& $94.56\%$& $234.29$ \\
        LEDNet \cite{wang2019lednet}& $61.25\%$& -  \\
        BiseNet \cite{yu2018bisenet}& $95.07\%$&  $74.21$\\
        DDRNet \cite{hong2021deep}& $95.51\%$&  $47.22$ \\
        \hline
        SPENet & $\pmb{95.88\%}$& \pmb{$27.09$}  \\
        \hline
        \end{tabular}
        \label{tab3}
        \end{center}
    \end{table}
\subsection{CMSE Results}
Consistency Mean Square Error(CMSE) stands as a pivotal indicator in affirming the efficacy of our proposed methodology. As can be observed from Table \ref{tab3}, our approach significantly outperforms any other model. The error value of $27.09\times10^{-4}$ of our approach is merely half that of DDRNet and ERFNet. In contrast, the performance of other models is notably inferior. Exemplar illustrations of segmentation results on the test set are depicted in Fig \ref{fig5}, through which we can observe that our model exhibits superior segmentation results, especially in the shape and consistency of the location hole.
\subsection{Ablation Study}
The inspiration for shape-aware comes from \cite{li2020improving} and based on it we propose the variable boundary domain(VBD). We explore the effectiveness of the Decoupled Module(DM) and VBD. DT denotes the distance from a pixel to the nearest pixel of another class that we choose as the threshold when generating the edge ground truth, DM denotes whether the decoupled module is utilized, VBD denotes whether the variable boundary domain is used. As shown in Table \ref{tab4}, In the case of using DM, mIoU has improved by at least $0.24\%$. When the DT increases from 1 to 3, the model's performance shows a slight decrease, but the difference is not significant. After incorporating VBD, the model's accuracy further improves, achieving an increase of $0.13\%$ compared to when DT is set to 1. To provide a more intuitive representation of our method, we generate heatmaps for both the boundary and body features shown in Fig \ref{fig6}.
    \begin{table}[htbp]
    \caption{ Ablative Experiments Results}
        \begin{center}
        \begin{tabular}{c|cccc}
        
        \cline{1-5} 
        \textbf{Model} & \textbf{DT}&\textbf{DM}& \textbf{VBD}& \textbf{mIoU}\\

        \hline
        SPENet &- &  &  & $95.44\%$\\
         & 1 & \checkmark & & $95.75\%$\\
         & 2 & \checkmark & & $95.73\%$\\
         & 3 & \checkmark & & $95.68\%$\\

         & - &  \checkmark& \checkmark & $\pmb{95.88\%}$\\
        \hline
        \end{tabular}
        \label{tab4}
        \end{center}
    \end{table}

\section{Conclusion}
In this paper, we design a novel lightweight network, which utilizes both body and edge information to improve the shape accuracy of segmentation while ensuring efficiency. A novel metric CMSE is proposed for describing segmentation consistency which holds great significance for industrial image analysis. Our method is designed for the specific PVC dataset but the performance on CityScapes is also competitive among numerous real-time networks. The variable boundary domain proposed in our study exhibits versatility and can be seamlessly integrated into other networks requiring boundary supervision. Moreover, our method can be transferred and applied to a wide range of visual tasks in industrial scenarios, thereby reducing labor costs and improving detection accuracy.

\bibliographystyle{ieeetr}
\bibliography{icme2023template}



\end{document}